\documentclass[10pt,twocolumn,letterpaper]{article}

\usepackage{cvpr}
\usepackage{times}
\usepackage{epsfig}
\usepackage{graphicx}
\usepackage{amsmath}
\usepackage{amssymb}

\usepackage{flushend}
\usepackage{subfigure}

\usepackage[pagebackref=true,breaklinks=true,letterpaper=true,colorlinks,bookmarks=false]{hyperref}

\cvprfinalcopy 

\def\cvprPaperID{****} 

\begin{document}
	
	\title{Automatic Whole-body Bone Age Assessment Using Deep Hierarchical Features}
	
	\author{Hai-Duong Nguyen\qquad Soo-Hyung Kim\\Chonnam National University\\Gwangju, South Korea\\{\tt\small nhduong\_3010@live.com}\qquad {\tt\small shkim@jnu.ac.kr}}
	
	
	\maketitle
	
	\begin{abstract}
		Bone age assessment gives us evidence to analyze the children growth status and the rejuvenation involved chronological and biological ages. All the previous works consider left-hand X-ray image of a child in their works. In this paper, we carry out a study on estimating human age using whole-body bone CT images and a novel convolutional neural network. Our model with additional connections shows an effective way to generate a massive number of vital features while reducing overfitting influence on small training data in the medical image analysis research area. A dataset and a comparison with common deep architectures will be provided for future research in this field.
	\end{abstract}
	
	
	\section{Introduction}\label{sec:introduction}
	
	Bone age is the indicator to determine the skeletal maturity of an individual. This index is the best way to assess the chronological ages by comparing with skeletal age. Hence, bone age assessment (BAA) is the common method to determine the final adult of the normal child or diagnosis of the children growth problem. Bone age has also used for determining the age at the condition where birth record and birth data are not available~\cite{mughal2014bone}. In addition, BAA can be used in the rejuvenation study associated with chronological and biological ages.
	
	All the previous studies merely consider evaluating the skeletal bone until the end of the adolescence when the form of the bone is completed. However, this leads to a question: how to cope with the maturity people. In such cases, BAA using left-hand X-ray does not work. Indeed, the image of bone such as CT scan of hand or full skeleton change according to the development and aging, from the whole bone level to the specific components~\cite{Boskey2010}. With aging, the bone becomes more fragile and less able to perform its mechanical functions. Hence, having a deep insight into skeleton change is a major interesting area of investigation, especially BAA of all age ranges. Considering that the bone morphology does not change when reaching adulthood, existing methods using left-hand X-ray image are not suitable for. Another approach is needed to overcome this issue.
	
	To this end, in this paper, we perform BAA by using whole-body CT scans. To the best of our knowledge, this is the first work of estimating the bone age for adults which also provides dataset for future research. Besides the contributions to the research of automated skeletal bone age analysis, this work will aim to build a novel deep learning based method for BAA by balancing the number of generated features in a network and avoiding overfitting with limited data. The outcome shows notable achievements in improving BAA accuracy based on whole-body CT image. Additionally, our research also shows the regions of interest (ROIs) impacting on BAA from the whole-body CT scan. Last but not least, we will demonstrate how gender plays a crucial role in BAA.
	
	The rest of the paper is organized as follows. In Section~\ref{sec:related_works}, we review conventional and deep learning based methods for BAA regarding their contributions and limitations. A novel BAA system based on whole-body CT scan will be introduced in Section~\ref{sec:architecture}. The major contribution of this work is to build a model that generates a massive number of features while reducing overfitting influence on small training data in the medical image analysis research area. In Section~\ref{sec:experimental_results}, the proposed model is evaluated and compared to other methods. This section also provides our findings on the relationship between BAA, bone locations (ROIs), and gender based on the experimental analysis. Section~\ref{sec:conclusion} concludes our study and comments future directions for further improvement.
	
	\section{Related Works}\label{sec:related_works}
	
	There are a variety of methods to compute the bone age. Over the past decade, most of the BAA had been developed using various visualization techniques. Those methods, for the most part, come from the X-ray or CT scan of the left-hand by comparing with the standard atlas of bone development based on a large number of collected images and assessed manually. G\&P is the most common method developed by Greulich and Pyle~\cite{greulich1959radiographic}. The radiologists used the entire reference images of the left wrist and hand which are standard for different ages of the individuals from birth till 19 years. This method is simple, but it can be influenced by the age population, race, and gender. Another renowned method is Tanner Whitehouse (TW)~\cite{tanner4goldstein,tanner1975assessment}. On the contrary, it depends on the set of specific ROIs in lieu of the whole hand. Each ROI have its own scores which will be summed up to give an overall bone maturity score. TW method is more complex and performs better than the G\&P method. Follow by TW approach, there are numerous other researches based on ROIs~\cite{4530646,5742275,5559448,6675918}. Although the outcome was improved, these traditional methods are still time-consuming and sensitive to the image quality.
	
	In addition to conventional methods such as G\&P or TW, recently, deep learning, particularly convolutional neural network (CNN), has rapidly become a methodology of choice for analyzing medical images~\cite{ litjens2017survey}. This method demonstrated performance improvements over the conventional methods on many problems in medical imaging. In this part, we review recent deep learning based methods that identify the age of a child from an X-ray of their hand. One of the well-known BAA systems using deep learning was proposed in the 2017 Pediatric Bone Age Challenge organized by Radiological Society of North America (RSNA)~\cite{rsna17}, the winner achieved the mean square error (MAE) of 4.265 months using multimodal technique considering both visual information and gender. Their backbone model was GoogLeNet combined with a simple multilayer perceptron taking human gender into account~\cite{rsna_winner}. That outcome exceeds the performance of conventional methods as well as some pediatric radiologists in an extensive experiment. Spampinato \etal~\cite{SPAMPINATO201741} provided a deep study of all possible cases by training from scratch, fine-tuning, and building a deep model named BoNet for BAA. Another work of Lee \etal~\cite{Lee2017} proposed a system that used LeNet for hand mask detection and GoogLeNet in age assessment. In the same manner, Iglovikov \etal~\cite{DBLP:journals/corr/abs-1712-05053} introduced a method that applied U-Net for hand segmentation, image registration based on key points, and a plain network inspired by VGGNet~\cite{Simonyan14c} for BAA. However, their results cannot lead to a better performance compared to the winner of the 2017 Pediatric Bone Age Challenge. Similarly, Zhou \etal~\cite{8227503} proposed a model based on VGGNet for BAA. Instead of using the entire X-ray image, they extracted ROIs then fed them into pre-trained models to obtain bone age by model fusion. As we stated above, unlike all the previous works which consider X-ray left-hand scans across up to 20 years of ages to monitor the growth status of the children, this paper introduces a novel network architecture for whole-body CT bone images in all age ranges that works well on small training data. In addition, source code and data will be also available to provide related materials for future research in this area.
	
	\section{Network Architecture}\label{sec:architecture}
	
	\begin{figure}[t]
		\begin{center}
			\includegraphics[scale=0.408]{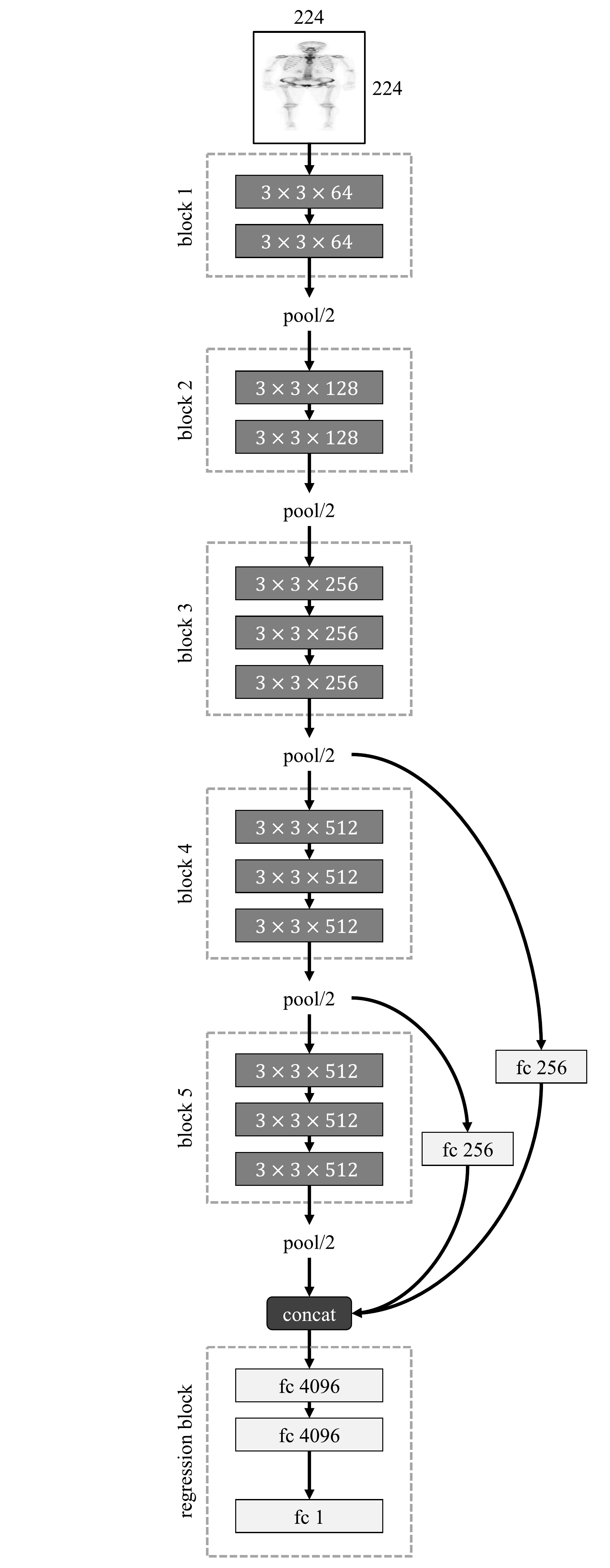}
		\end{center}
		\caption{Whole-body CT bone age estimation model. Both local and global information are considered for regression which generate a massive number of features for small medical training data.}
		\label{fig:mlvgg}
	\end{figure}
	
	In this section, we provide the details of networks for automatic BAA using full-body CT images. Various architectures and learning techniques were taken into account including: fine-tuning models originally trained on the large-scale ImageNet dataset, ensembling fine-tuned models, and building a new model for BAA.
	
	\subsection{Fine-tuned Models}\label{sec:sub_fine_tuned}
	
	CNNs have achieved many successes in image classification owning to the availability of public large-scale datasets and high performance GPUs. VGGNet, GoogLeNet~\cite{7298594}, and ResNet~\cite{7780459} are excellent examples that outperformed the others in the ImageNet Challenge and they are still useful for transfer learning to tackle related computer vision issues.
	
	In general, given an image, deep neural networks automatically generate simple, moderate, and complex information via low-level, mid-level, and high-level layers respectively. Even though the extracted features from these models can be used directly for a particular task, the patterns are significantly different in medical images. Therefore, a great amount of annotated data are required to train network layers. However, this requirement is not always reserved especially in medical image analysis research area. Hence, reusing first layers of a well-trained model and fine-tuning last layers is a straightforward solution to this issue.
	
	In this paper, we fine-tune VGGNet, GoogLeNet, and ResNet on our dataset for automatic BAA. In other words, mid-level and high-level layers in the base architectures and a regression network will be trained to be adapted to our data while the early layers in the backbone models are reserved. Training configurations will be discussed in Section \ref{sec:sub_implementation} and the best fine-tuned models will be use for ensembling.
	
	\subsection{Hierarchical Features for Whole-body BAA}\label{sec:sub_mlvgg}
	
	In this section, we introduce a model which is based on VGGNet for automatic BAA. By combining features from different levels of the base network, our model generates a large amount of essential information related to whole-body bone age. Moreover, trivial features are also eliminated from the early stages by simple fully connected layers.
	
	Our base model is a VGGNet which is a conventional CNN with stacks of convolutional layers arranged in 5 blocks. Unlike deep architectures such as GoogLeNet or ResNet which take advantage of additional network connections to save activations from dying, the role of the added connections in our model as shown in Fig.~\ref{fig:mlvgg} is to enlarge number of features for our limited training data. In place of taking only high-level features into account, our model considers information from different layers of the base architecture for regression. That is, both global and local features will be used in a single network.
	
	\begin{figure}[t]
		\begin{center}
			\includegraphics[scale=0.408]{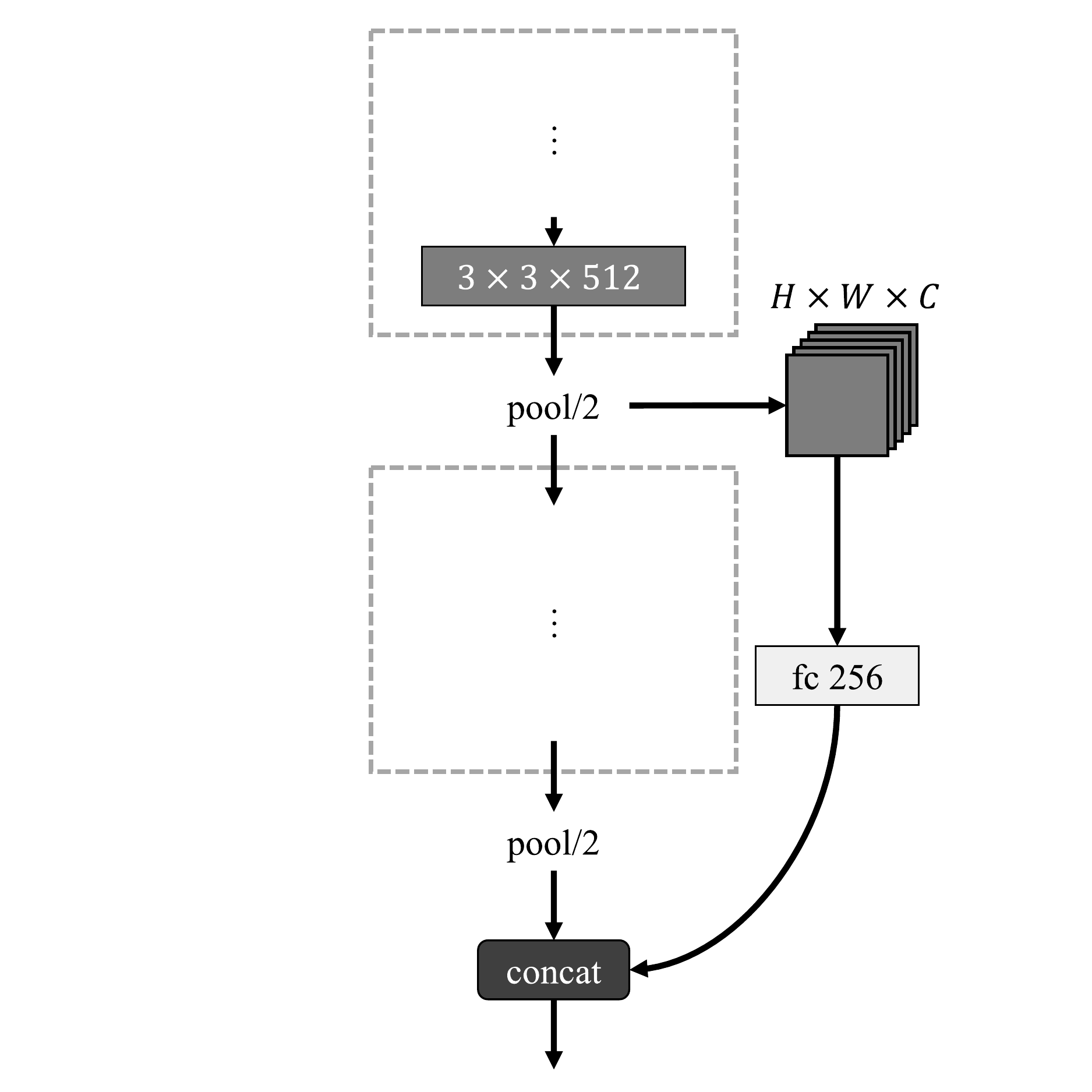}
		\end{center}
		\caption{An additional connection uses local features in our model.}
		\label{fig:add_connection}
	\end{figure}
	
	Let $C$ be the number of feature maps generated by a specific layer of the network, $H$ and $W$ be the height and width of each feature map respectively (Fig.~\ref{fig:add_connection}), then the number of learnable parameters of an additional connection in our model is given as follow
	\begin{equation}
	N(C,H,W,n_U)=C\times H\times W\times n_U + n_U ,
	\end{equation}
	where $n_U$ is the number of units of the fully connected layer in the connection. It can be seen that the feature map size ($H\times W$) at the low-level layer is not small regularly. In addition, they are supposed to contain elementary features which are trivial for the final task of the network. Therefore, to avoid complicating the model, the information from the first two blocks are not directly used for regression. Furthermore, since a small number of mid-level features play an important role in the final regression, we insert a fully connected layer with $n_U=256$ units in network connections. The main purpose of these layers is to reserve useful information while insignificant features will be eliminated. Then we concatenate all extracted features and feed them into a regression network containing three fully connected layers. By adding two simple network connections, our model outperforms other architectures in the comparison with the same data and training configurations. Since the network contains several intermediate layers, we may have other promising choices to gather mid-level features for regression. However, as increasing number of additional connections, the architecture will be more complex. We chose the best model including two connections with max-pooling layers from block \#3 and block \#4. It is reasonable to assume that the abstracted features generated by max-poolings are more effective compared to the others.
	
	\section{Experimental Results}\label{sec:experimental_results}
	
	This section describes the dataset and metric that we used to evaluate the models. The details of network implementation and a performance comparison between architectures will be covered also. Last but not least, an experimental analysis will be discussed to provide some interesting findings related to age estimation using full-body bone CT scan.
	
	\subsection{Dataset}\label{sec:sub_dataset}
\begin{figure}[t]
		\begin{center}
			\subfigure[6 years old\newline \hspace*{1.3em} male]
			{
				\includegraphics[width=0.7in]{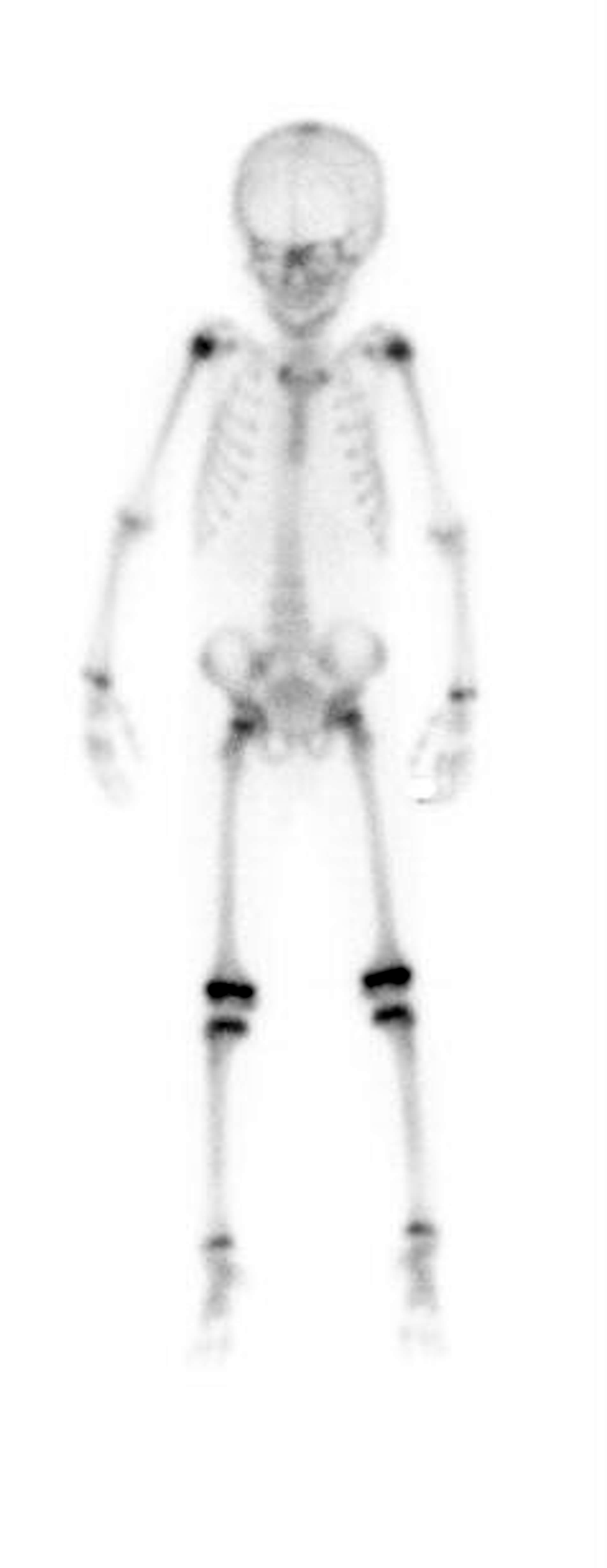}
				\label{fig:sub_sample_1}
			}
			\subfigure[18 years old\newline \hspace*{1.3em} male]
			{
				\includegraphics[width=0.7in]{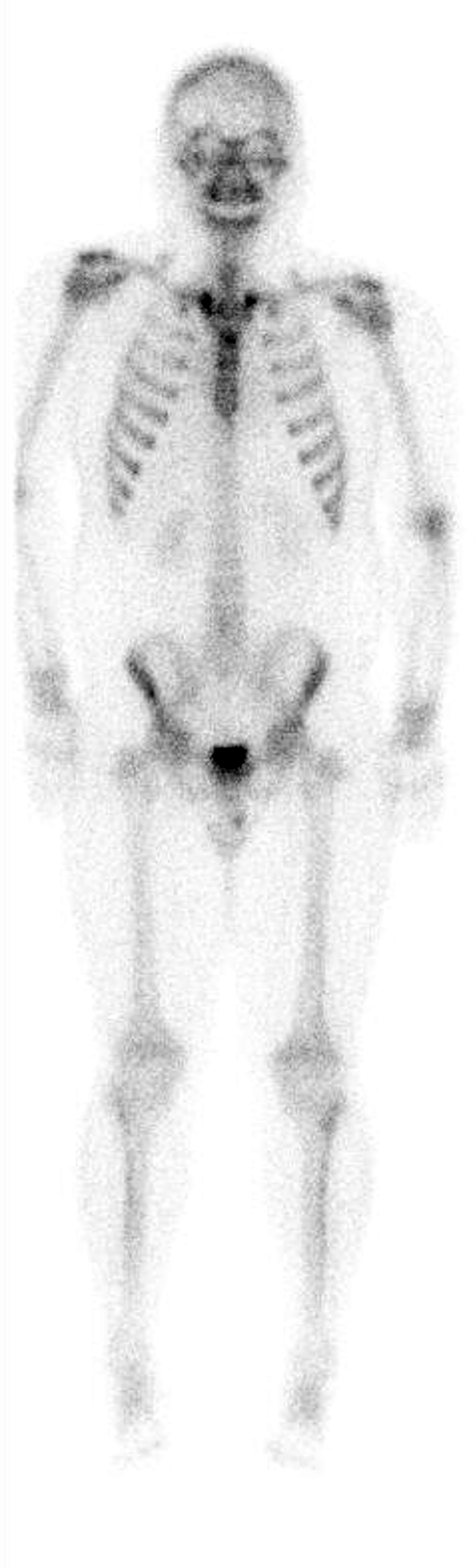}
				\label{fig:sub_sample_2}
			}
			\subfigure[49 years old\newline \hspace*{1.3em} female]
			{
				\includegraphics[width=0.7in]{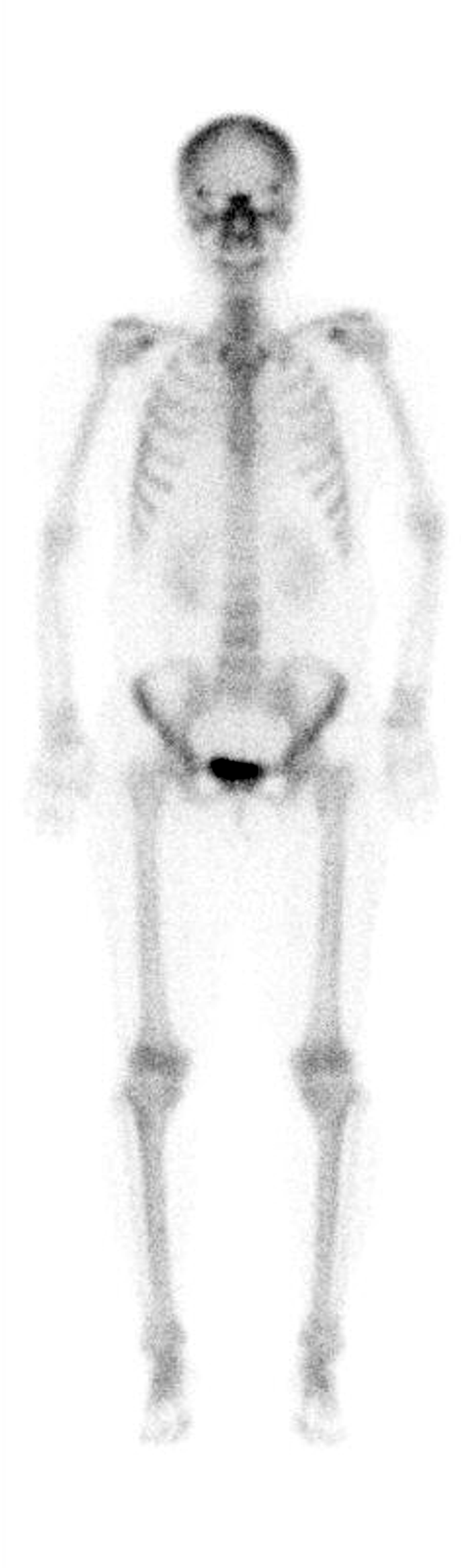}
				\label{fig:sub_sample_3}
			}
			\subfigure[74 years old\newline \hspace*{1.3em} male]
			{
				\includegraphics[width=0.7in]{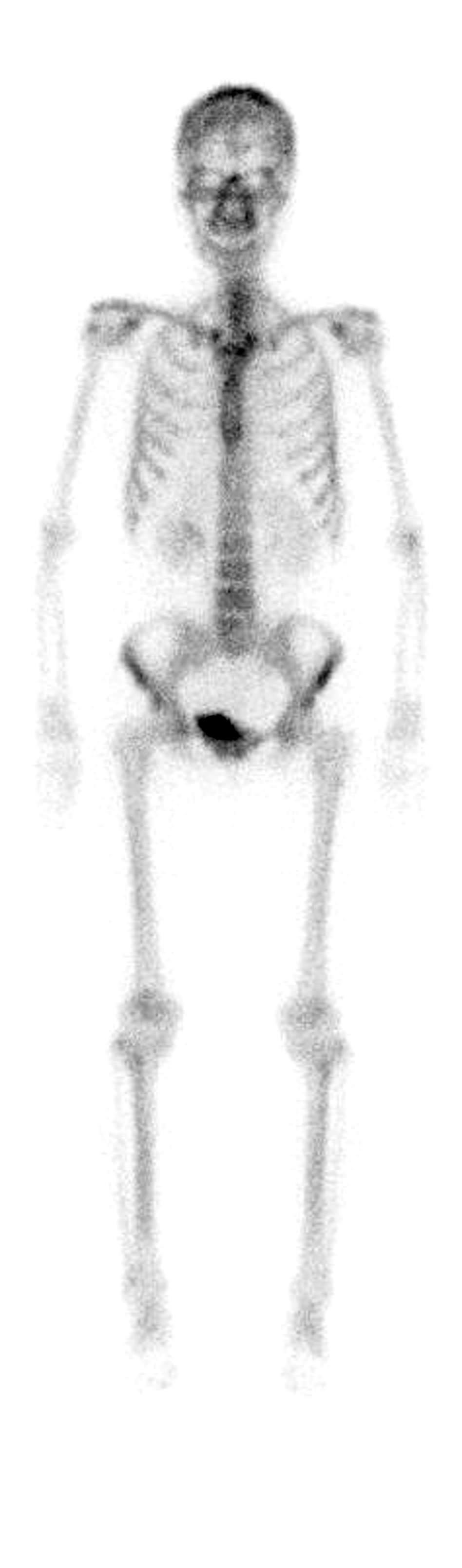}
				\label{fig:sub_sample_4}
			}
		\end{center}
		\caption{Whole-body CT bone scan samples with corresponding age and gender.}
		\label{fig:samples}
	\end{figure}

\begin{figure}[t]
		\begin{center}
			\subfigure[Data distribution w.r.t. age]
			{
				\includegraphics[height=2.0in]{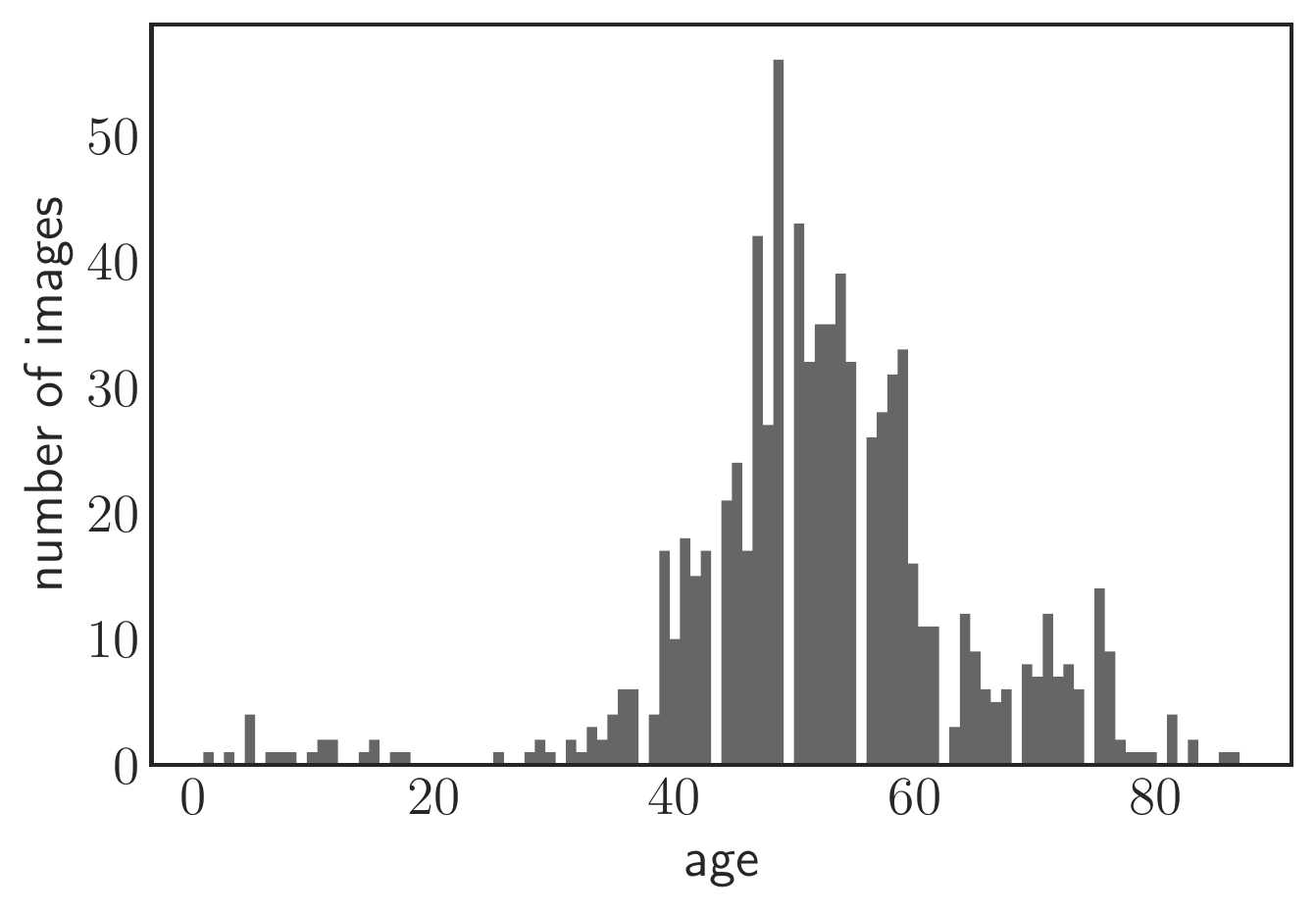}
				\label{fig:sub_data_dis_age}
			}\\
			\subfigure[Data distribution w.r.t. gender]
			{
				\includegraphics[height=1.7in]{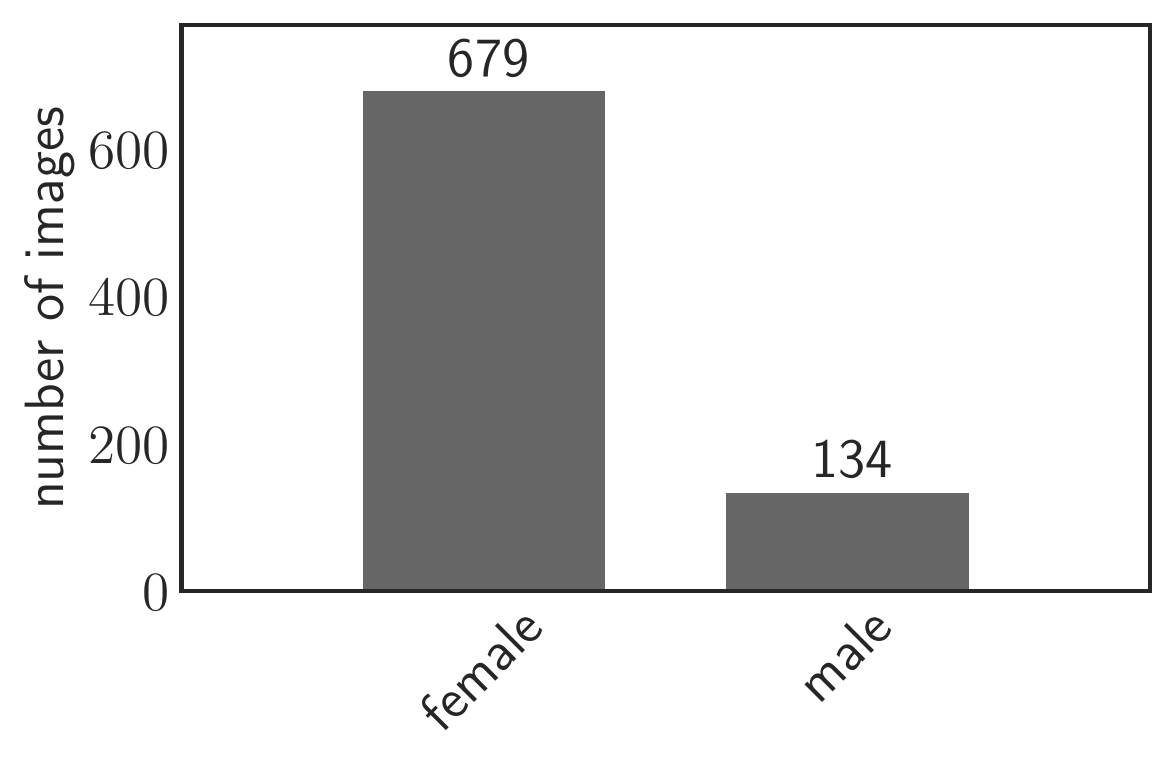}
				\label{fig:sub_data_dis_gen}
			}
		\end{center}
		\caption{Our whole-body CT bone scan data distribution.}
		\label{fig:data_dis}
	\end{figure}

	In the previous works, X-ray left-hand scans of children of the age up to 18 years old are considered. On the other hand, our dataset was provided by \textit{Anonymous} hospital containing $813$ whole-body CT bone images with the age range between $8$ months and $87$ years. All the images are grayscale and the average resolution is $3000\times900$ pixels. Each image was annotated with corresponding age (in years) and gender (see Fig.~\ref{fig:samples} for examples). Fig.~\ref{fig:sub_data_dis_age} shows the data distribution with respect to age while Table~\ref{fig:sub_data_dis_gen} represents number of data for each gender. $73.432\%$ of the data are at the age from $40$ to $60$ while $8.487\%$ and $18.081\%$ of the data belong to the age ranges of $0-39$ and $61-87$, respectively. In addition, it is noticeable that we have a serious imbalanced data problem in terms of gender. Even though the purpose of this research is to estimate age via whole-body CT scan, we will discuss the influence of gender on BAA in the next section. As the number of training images is limited, it provides an interesting benchmarking platform for different learning methods, particularly deep neural networks which acquire a reputation for the performance on large-scale datasets. We used this dataset in our experiments by randomly dividing it into two partitions: $569$ images for training ($70\%$) and $244$ images for testing ($30\%$).
	
	\subsection{Implementation}\label{sec:sub_implementation}
	
	All the models were implemented using PyTorch~\cite{paszke2017automatic}. Training images were rescaled to $256\times256$ and $331\times331$ before random cropping to $224\times224$ and $299\times299$ adapting to common architectures VGGNet, ResNet, and GoogLeNet in our comparison. We also normalized the data to have zero mean and unitary standard deviation following the practice in~\cite{SPAMPINATO201741}. Random horizontal flipping is used also. After each convolutional layer and before ReLU nonlinearity, we applied batch normalization~\cite{Ioffe:2015:BNA:3045118.3045167} to improve the network’s generalization capability. In addition, dropout~\cite{DBLP:journals/corr/abs-1207-0580} was used after each fully connected layer to serve the same purpose.
	
	VGGNet, GoogLeNet, and ResNet were initialized with weights on ImageNet dataset then fine-tuned on our training data. Originally, these models aim to recognize 1000 classes using softmax classifier as the last layer of the network. In this work, we replace softmax classifiers with single unit linear layers for age estimation. In case of our model for BAA, we initialized the backbone network using original VGGNet while two additional connection layers' weights were generated randomly. Then these layers were trained from scratch and the base model was fine-tuned on the same training data as described above.
	
	All the models were used pre-trained models as the initial conditions, which massively speed up the training procedure. In addition, this technique is remarkably effective to prevent overfitting on our small training data. In our experiments, the models were fine-tuned on the same data for up to 130 iterations. The weights were updated by Adam~\cite{adam2014} with a minibatch size of 256. The learning rate started at $3.10^{-4}$ and the minimum value was $10^{-7}$. During the training procedure, the learning rate will be reduced by the factor of 0.8 after 10 iterations with no improvement in validation loss. To avoid overfitting, online data augmentation was performed on the training data. Data preprocessing techniques comprised normalization, random cropping, and mirroring.
	
	\subsection{Experimental Analysis}\label{sec:sub_analysis}
	
	This section reports the performances achieved by our models in terms of MAE. Comparisons will be provided and we will show how our model outperforms the others. Furthermore, we visualize the ROIs that we should pay attention to estimate age using whole-body CT images. An assumption about the relationship between bone age and gender will be made based on the experiments also. 
	
	In the first comparison, we take the fine-tuned VGGNet and our modified VGGNet into account. The difference is that our model takes advantage of local features in intermediate layers of the VGGNet to generate better visual representations as shown in Fig.~\ref{fig:mlvgg}. Table~\ref{tab:vgg_vs_mlvgg} shows a significant improvement thanks to our additional network connections on the same data and learning configurations. In place of training the network from scratch on a small dataset which causes overfitting, we use local features generated from well-learned intermediate layers of VGGNet. Then, these mid-level information will be combined with global features that adapt to our data by training. Fully connected layers are added inside the connections and after concatenation layer to minimize the number of trivial information used for regression. It is noticed that our model can still be trained end-to-end using available optimization algorithms such as SGD, Adam, etc..
	
	\begin{table}
		\begin{center}
			\begin{tabular}{|l|c|}
				\hline
				Model                          & \begin{tabular}[c]{@{}c@{}}MAE\\ (years)\end{tabular} \\ \hline \hline
				Fine-tuned VGGNet              & 9.741                                                    \\
				Our modified VGGNet & 4.856
				\\ \hline
			\end{tabular}
		\end{center}
		\caption{Performance comparison between the fine-tuned VGGNet and our modified VGGNet with respect to MAE on testing data.}
		\label{tab:vgg_vs_mlvgg}
	\end{table}
	
	We showed how additional connections improve a VGGNet in BAA. In the next comparison, we will take common deep architectures including GoogLeNet, ResNet, and ensemble models into account. In our experiments, all the models' weights were updated using the same data and training settings. Each model was trained several times and the best performances were picked. See Table~\ref{tab:mlvgg_vs_all} for the comparison.
	
	\begin{table}
		\begin{center}
			\begin{tabular}{|l|c|}
				\hline
				Model                                   & \begin{tabular}[c]{@{}c@{}}MAE\\ (years)\end{tabular} \\ \hline \hline
				Fine-tuned VGGNet (A)                   & 9.741                                                 \\
				Fine-tuned GoogLeNet (B)             & 5.522                                                 \\
				Fine-tuned ResNet (C)                   & 5.738                                                 \\
				Early fusion of (A), (B), and (C)       & 6.870                                                  \\
				Late fusion of (A), (B), and (C)        & 5.140                                                  \\
				\textbf{Our VGGNet with hierarchical features} & \textbf{4.856}                                        \\ \hline
			\end{tabular}
		\end{center}
		\caption{Performance comparison between our model and common architectures for BAA in terms of MAE.}
		\label{tab:mlvgg_vs_all}
	\end{table}
	
	Among the popular deep architectures, GoogLeNet obtained the best performance for BAA with the MAE of $5.522$. We argue the reason is that GoogLeNet with \textit{Inception} modules considering different sizes of the kernel are better at understanding local features in a bone age image. This finding might be discovered by Cicero \etal~\cite{rsna_winner} in the 2017 RSNA ML Challenge as well. As they stated, many architectures had been tested and GoogLeNet was the best one to be the winner of the challenge. In our experiments, the performance can be improved by the ensemble of chosen models. We did the fusion in two scenarios: early fusion and late fusion. Although the former can bring the benefit of gathering a variety of features from different type of architectures, the ensemble model seemed to suffer from overfitting due to our small training data compared to the number of learnable parameters of the entire model. On the other hand, late fusion by averaging the ages computed by three models leads to noticeably accurate outcomes with MAE of $5.140$. In our comparison, the best result was achieved by our modified VGGNet with two additional connections. It learned a large number of features from small data and performed better than other single and ensemble models. The key point is that these additional network connections can be easily attached to an arbitrary model. It is still reasonable although the network complexity should be increased.
	
\begin{figure}[t]
		\begin{center}
			\subfigure[Whole-body CT bone images]
			{
				\includegraphics[width=1.0in]{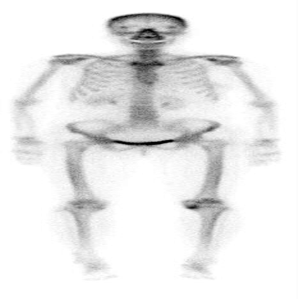}
				\includegraphics[width=1.0in]{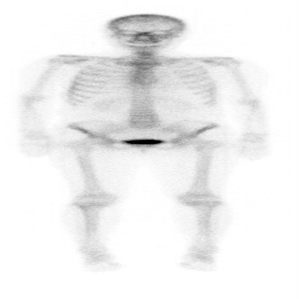}
				\includegraphics[width=1.0in]{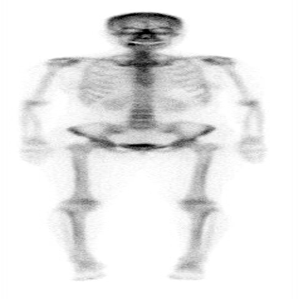}
				\label{fig:sub_vital_reg_11}
			}\\
			\subfigure[Grad-CAMs]
			{
				\includegraphics[width=1.0in]{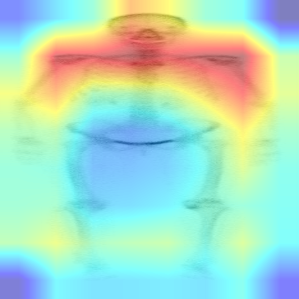}
				\includegraphics[width=1.0in]{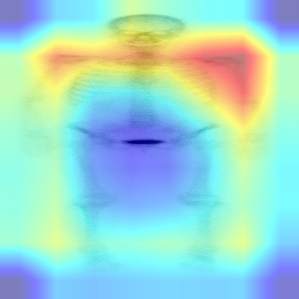}
				\includegraphics[width=1.0in]{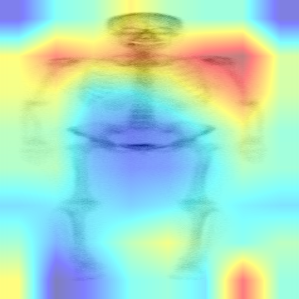}
				\label{fig:sub_vital_reg_21}
			}
		\end{center}
		\caption{ROIs in whole-body CT bone images for BAA produced by our modified VGGNet and Grad-CAM.}
		\label{fig:vital_regions}
	\end{figure}

	In addition to deep architecture comparison, in this section, we provide an interesting insight into ROIs via Grad-CAM~\cite{Visual} that may play a critical role in manual BAA. Fig.~\ref{fig:vital_regions} shows whole-body CT bone images and their corresponding Grad-CAMs indicating the ROIs that the model tends to focus on for age estimation.
	Interestingly, the upper body comprises several characteristics related to age assessment. To verify this perception, we carried out an experiment by retraining modified VGGNets in two scenarios. The former takes an upper body image as the input while the latter considers a lower body CT scan instead. Table~\ref{tab:data_type_res} clarifies the Grad-CAMs in Fig.~\ref{fig:vital_regions} as the upper body plays the main role in BAA using our model.
	
	\begin{table}[]
		\begin{center}
			\begin{tabular}{|l|c|}
				\hline
				Data type  & \begin{tabular}[c]{@{}c@{}}MAE\\ (years)\end{tabular} \\ \hline \hline
				Upper body & 4.854                                                 \\
				Lower Body & 5.100                                                 \\ \hline
			\end{tabular}
		\end{center}
		\caption{The influence of ROIs in BAA.}
		\label{tab:data_type_res}
	\end{table}
	
	Furthermore, we also inspected the relationship between gender and bone structure for automatic BAA. As we mentioned before, over $83\%$ of patients in our dataset are females. Since our architecture does not consider gender as an attribute, the number of data in each gender should not influence the final performance. Conversely, our experiments show that the differences between male and female human skeleton structure may be an important characteristic for BAA. With a larger number of female instances, the model was likely to discover specific properties in female bone that do not hold in male skeleton (see Table~\ref{tab:gender_res}).
	
	\begin{table}[]
		\begin{center}
			\begin{tabular}{|l|c|c|}
				\hline
				Gender & Number of data & \begin{tabular}[c]{@{}c@{}}MAE\\ (years)\end{tabular} \\ \hline \hline
				Female & 679            & 4.351                                                 \\
				Male   & 134            & 6.969                                                 \\ \hline
			\end{tabular}
		\end{center}
		\caption{Gender influence in BAA.}
		\label{tab:gender_res}
	\end{table}
	
	\section{Conclusion}\label{sec:conclusion}
	
	This paper addressed the whole-body CT scan BAA using hierarchical features from CNN. We tested several pre-trained models on our small dataset before building a new model based on VGGNet for automatic BAA. Our model with additional connections aims to balance the number of generated features in a network and avoiding overfitting with limited data in the medical image analysis research area. In addition, we also provided insights into the ROIs and the influence of gender in BAA. These interesting findings and our public dataset may supply an useful benchmarking platform for further research in whole-body CT image BAA. To the best of our knowledge, this is the first work in bone age estimation using a whole-body skeleton CT scan which also provides related materials for future investigation. Since our network connections may complicate the entire architecture, a global pooling layer~\cite{glob_pool} can be attached right before the connections to reduce the number of network parameters.
	
	{\small
		\bibliographystyle{ieee}
		\bibliography{egbib}
	}
	
\end{document}